\documentclass{isprs} 
\usepackage{subfigure}
\usepackage{setspace}
\usepackage{geometry} 
\usepackage{epstopdf}
\usepackage[labelsep=period]{caption}  
\usepackage[british]{babel} 
\usepackage[hang]{footmisc}
\usepackage{float}

\usepackage{tabularx}
 \usepackage{booktabs}
 \usepackage{natbib} 
 \usepackage{siunitx}
 \usepackage{varwidth}
\usepackage{amsmath}
\usepackage{xcolor}
\usepackage{float}
\usepackage{tabularx,booktabs}
\usepackage{siunitx}
\usepackage{subfigure} 
\usepackage{url}
\usepackage{subfigure}

\begin{document}





\title{A Comparative Neural Radiance Field (NeRF) 3D Analysis of Camera Poses from HoloLens Trajectories and Structure from Motion}
\author{
 Miriam Jäger\thanks{Corresponding author}, Patrick Hübner, Dennis Haitz, Boris Jutzi}

\address{
	Institute of Photogrammetry and Remote Sensing (IPF)\\ Karlsruhe Institute of Technology (KIT), Karlsruhe, Germany\\
	
	(miriam.jaeger, patrick.huebner, dennis.haitz, boris.jutzi)@kit.edu\\
}



\abstract{
Neural Radiance Fields (NeRFs) take a set of camera poses with associated images as input and train a Neural Network which estimates a density value and color values for each position. The position-dependent learning of density is of particular interest for photogrammetry. The coordinate system of the trained NeRF can be retrieved and filtered according to its density on objects, which enables a 3D reconstruction.
Most commonly, traditional methods like Structure from Motion (SfM) are used to calculate the camera poses in pre-processing needed for training NeRFs.
From this perspective, the HoloLens provides an interesting interface, that enables the extraction of the required input data for NeRFs, the camera poses and associated sensor RGB images. In this work we presented a workflow for the extraction of high resolution 3D reconstructions almost directly from HoloLens data under the application of NeRFs. 
On the one hand, internal camera poses retrieved from the HoloLens trajectory via a server application, On the other hand, camera poses via Structure from Motion were considered. For both camera poses, an enhanced variant was additionally applied by a pose refinement while training.
The results show that the internal camera poses lead to convergence of the NeRF after a simple rotation around the x-axis and enable a 3D reconstruction.
Improvements in training and reconstruction can be achieved by pose refinement while training. This enables a comparable quality in the training process and resulting point cloud as achieved by external camera poses in pre-processing caculated using Structure from Motion.
In general all NeRF reconstructions outperform the conventional photogrammetric method using Multi-View Stereo.
}

\abstract{Neural Radiance Fields (NeRFs) are trained using a set of camera poses and associated images as input to estimate density and color values for each position. The position-dependent density learning is of particular interest for photogrammetry, enabling 3D reconstruction by querying and filtering the NeRF coordinate system based on the object density. While traditional methods like Structure from Motion are commonly used for camera pose calculation in pre-processing for NeRFs, the HoloLens offers an interesting interface for extracting the required input data directly. We present a workflow for high-resolution 3D reconstructions almost directly from HoloLens data using NeRFs. Thereby, different investigations are considered: Internal camera poses from the HoloLens trajectory via a server application, and external camera poses from Structure from Motion, both with an enhanced variant applied through pose refinement. Results show that the internal camera poses lead to NeRF convergence with a PSNR of 25\,dB with a simple rotation around the x-axis and enable a 3D reconstruction. Pose refinement enables comparable quality compared to external camera poses, resulting in improved training process with a PSNR of 27\,dB and a better 3D reconstruction. Overall, NeRF reconstructions outperform the conventional photogrammetric dense reconstruction using Multi-View Stereo in terms of completeness and level of detail.}

\maketitle

\section{INTRODUCTION}\label{INTRODUCTION}
 
\sloppy


With the pioneering research on Neural Radiance Fields (NeRFs) \citep{mildenhall_et_al_2020}, that enable the rendering of new views with the so-called view synthesis based on image data and associated camera poses in space, a new epoch in computer graphics started. Novel inventions like Instant NGP \citep{mueller_et_al_2022} advanced NeRFs once again, as they reduce training and rendering time to minutes or even seconds.\newline

However, these methods also arouse interest beyond the field of computer graphics for research and development in photogrammetry. 
NeRFs take a sparse set of camera poses with associated images as input and train a Neural Network which estimates a density value $\delta$ and color values c = (R,G,B) for each position X = (x,y,z).
The position-dependent learning of density, and color values, is of particular interest for photogrammetry with regard to mobile 3D mapping. We consider density as a kind of pseudo-probability for the occurrence of an object in 3D space. Thus, the positions of the trained NeRF can be accessed and filtered by their density on objects, which allows the extraction of 3D point clouds of the scene. \newline 

Most commonly, traditional methods like Structure from Motion (SfM) are used to calculate the camera poses in pre-processing needed for training the NeRFs.
From this perspective, the HoloLens provides an interesting interface, that enables the extraction of the required input data, the camera poses and associated sensor RGB images. This allows to create a 3D reconstruction, including color information, nearly directly from the sensor data. 
\newpage
In this study, we investigate whether the trajectory from the HoloLens is sufficient to achieve convergence of the NeRF and the potential to achieve a 3D reconstruction of the scene based on the density values of the NeRF trained with HoloLens data. Four different types of camera poses using the same HoloLens images are compared regarding their training process and the resulting 3D reconstructions. On the one hand, the internal HoloLens camera poses, as well as a including pose refinement during training are considered. On the other hand, external generated camera poses via SfM, as well as including the pose refinement are investigated.
In order to compare the point clouds based on NeRFs with traditional methods, a dense Multi-View Stereo (MVS) point cloud from the camera poses is reconstructed.\newline

Firstly, it was demonstrated that the internal HoloLens camera poses and images are suitable for the convergence of the NeRF, as shown by the quantitative results of the training process in Figure \ref{fig:training}. After a simple rotation around the x-axis, the convergence occurs from approximately 20,000 training epochs with a Peak-Signal-to-Noise-Ratio (PSNR) of 25\,dB. 
Secondly, the trained NeRF is suitable for a 3D reconstruction of the scene, as shown by the qualitative results in Figure \ref{fig:pointclouds_COLMAP_SLAM}.
After additional training of extrinsics alias pose refinement, HoloLens camera poses lead to comparable PSNR values of 27\,dB and a comparable 3D reconstruction with respect to the separate pose determination via SfM in pre-processing. 
Furthermore, the reconstruction from the NeRF exhibited advantages over a conventional dense 3D reconstruction through MVS. The reconstructions from the NeRF using SfM and internal HoloLens camera poses lead to a higher point density, less artefacts, and a better mapping of untextured surfaces in terms of completeness than MVS.

\section{RELATED WORK}\label{RELATED WORK}
In this section, we briefly summarize related work
to our study. Thereby we give an overview on basic and recent research and development on NeRFs. \newline

The foundation for the Neural Radiance Fields (NeRFs) was established by the Scene Representation Networks (SNR) \citep{SceneRepresenationNetworks}. Their underlying principle is modeling the scene as a function of 3D coordinates within it. It was followed by the groundbreaking research work of Neural Radiance Fields \citep{mildenhall_et_al_2020}. These enable estimation of color values and densities for each 3D coordinate through 6D camera poses and associated 2D images by learning a deep neural network with multi-layer perceptrons (MLPs). The initial NeRF was followed by thousands of publications driving research and development in various domains. \newline

To address scalability, the augmentation to large scale scenes is achieved by Mega-NeRF \citep{Mega-nerf} using a data partitioning based on visibility analysis or Block-NeRF \citep{Block-nerf} with distance-dependent partitioning based on street segments.
Other approaches, such as Bundle Adjusting Radiance Fields (BaRF) \citep{Barf} and Gaussian Activated Radiance Fields (GaRF) \citep{Garf}, address the purpose of a camera pose estimation.
In addition to neural methods, non-neural research like Plenoxels \citep{plenoxels} have been proposed. Dynamic contributions, such as \citep{D_nerf}, utilize time as an additional input dimension for time-dependent rendering of novel images, while \citep{dynamic_nerf} employ time components for preventing the occurrence of artifacts caused by dynamic pixels.
Furthermore, 3D reconstruction from NeRFs are considered \citep{Slam_nerf, unisurf}.
Methods such as AdaNeRF \citep{adanerf}, FastNeRF \citep{fastnerf} and Instant NGP \citep{mueller_et_al_2022} aim to improve rendering or training efficiency. Thereby Instant NGP, which we utilize in our study, uses a combination of small MLPs and spatial hash table encoding for real-time training and rendering.







\section{METHODOLOGY}\label{sec:METHOD}
\begin{figure}[h]
\begin{center}
		\includegraphics[width=1\columnwidth]{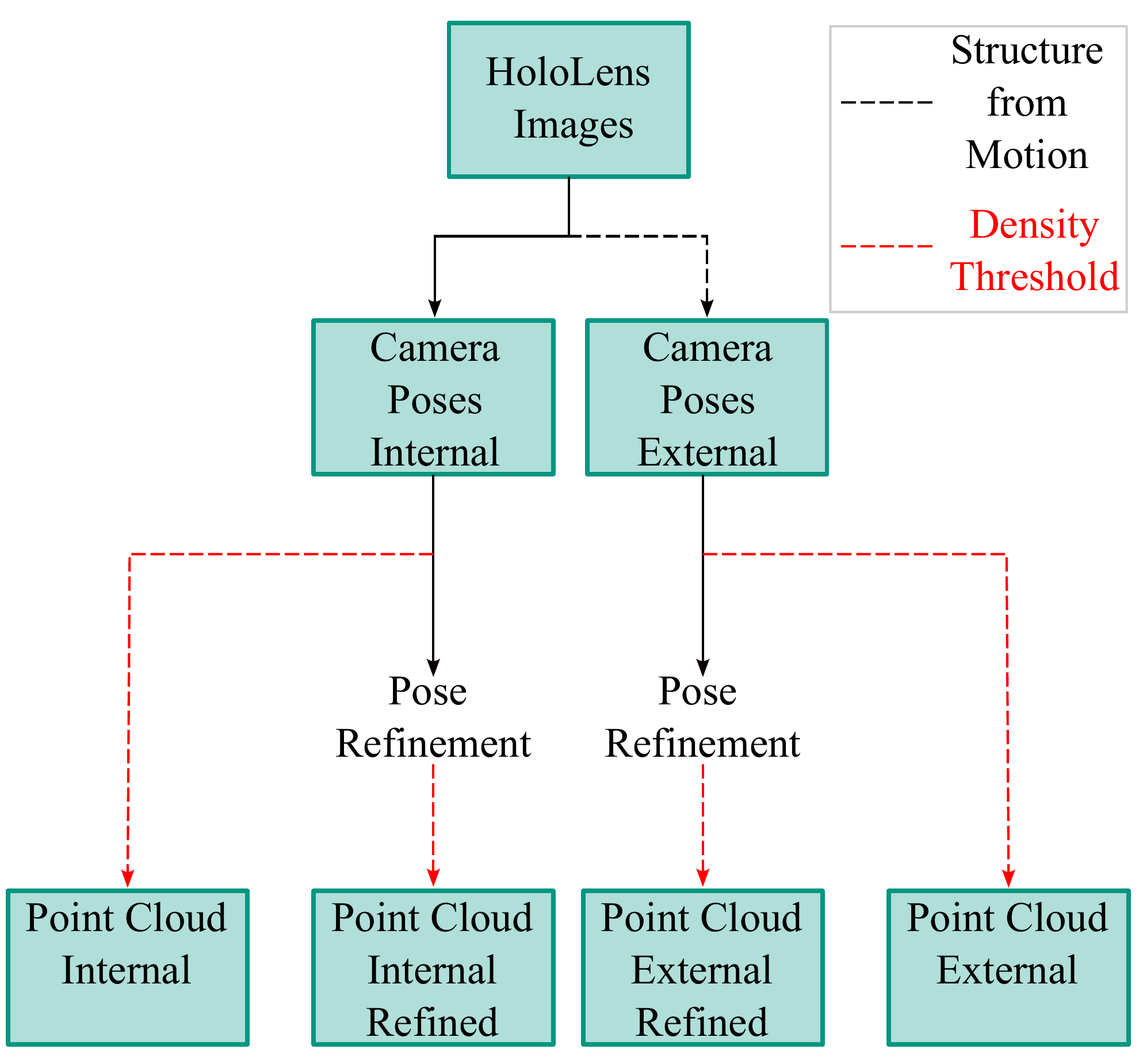}
	\vspace{-3mm}
	\caption{Flowchart of the applied investigations. Input data are two different types of camera poses: internal poses of the HoloLens and externally calculated poses via SfM. Subsequently, in each case a pose refinement variant is performed during the training. The four resulting point clouds are extracted from the NeRF using a global density threshold.}
\label{fig:flowchart}
\end{center}
\end{figure}

In Section \ref{sec:pose_transform} the principles of the methods used to generate the input data according to Figure \ref{fig:flowchart} for the NeRFs are presented. The essential transformations of the internal HoloLens camera poses in order to use them for the NeRFs are explained. After that, the standard method used to determine the camera poses in pre-processing is introduced. Subsequently, in Section \ref{sec:reconstruction_method} the methodology for the extraction of a 3D point cloud from NeRFs as well as the used conventional photogrammetric reconstruction are described.

\subsection{Camera Poses of the Trajectories}\label{sec:pose_transform}
\begin{figure*}[h!]
\centering
\subfigure[]{\label{fig:01_pose_original}\includegraphics[width=0.23\textwidth]{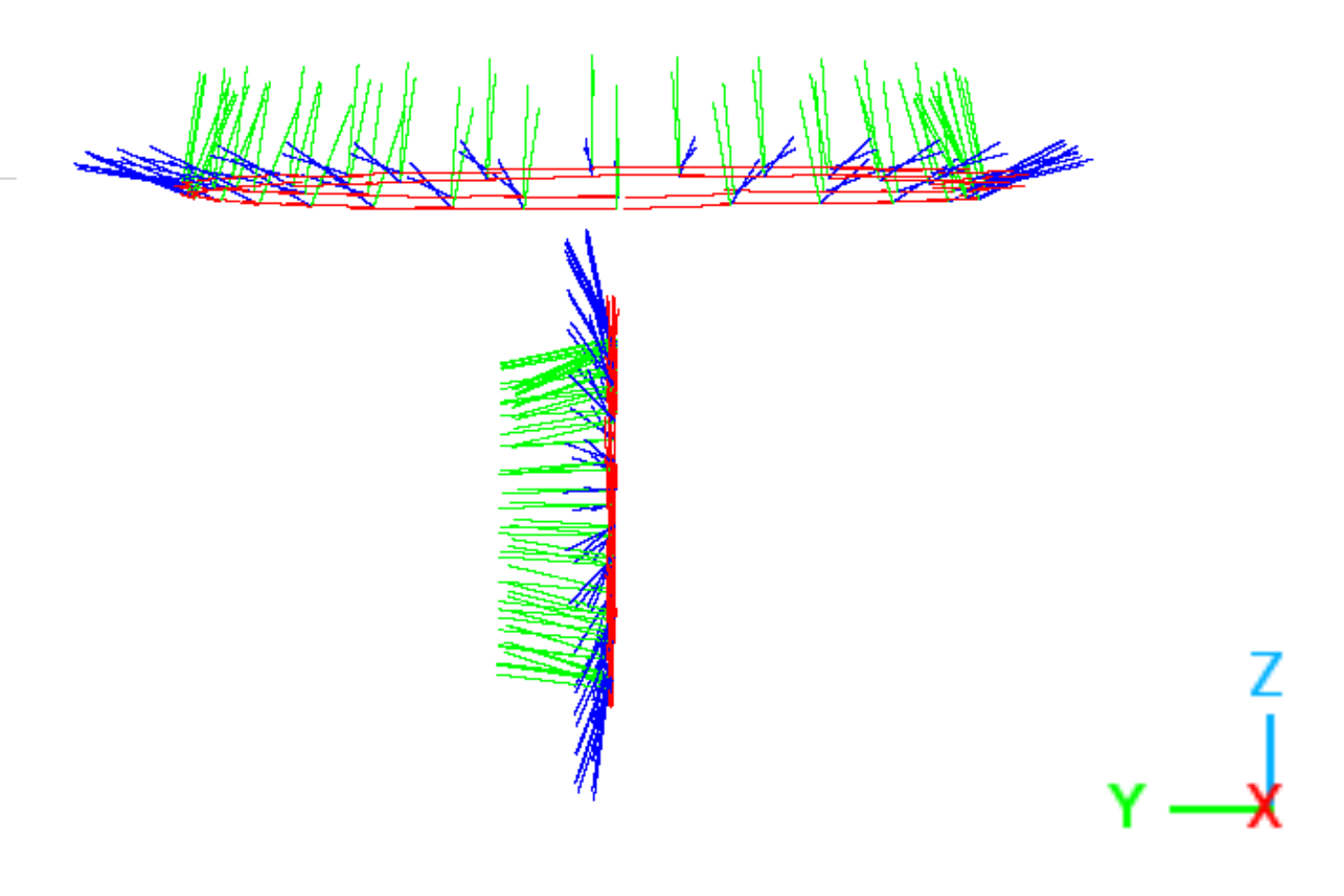}}
\subfigure[]{\label{fig:02_pose_x-achse}\includegraphics[width=0.23\textwidth]{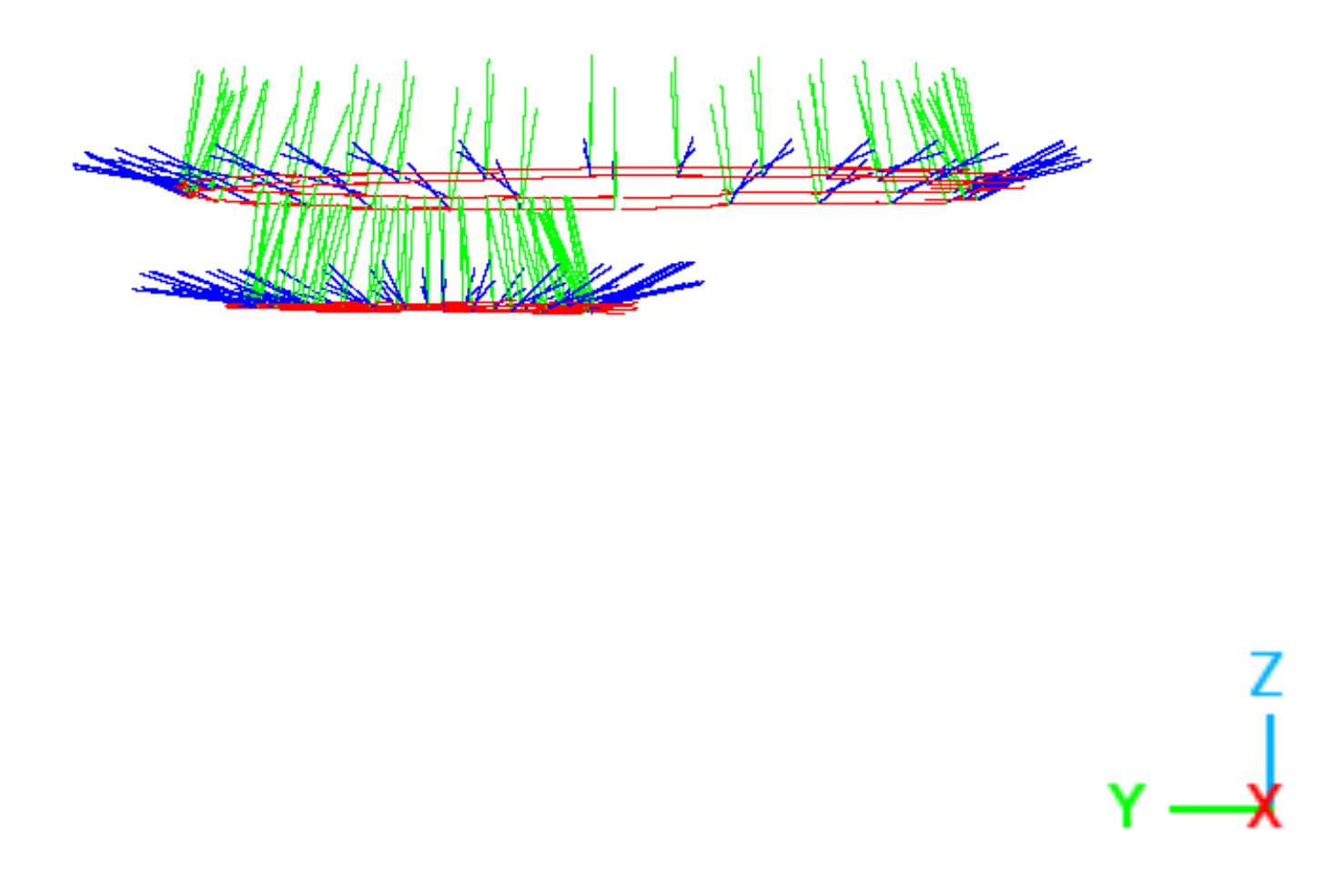}}
\subfigure[]{\label{fig:03_pose_attention}\includegraphics[width=0.23\textwidth]{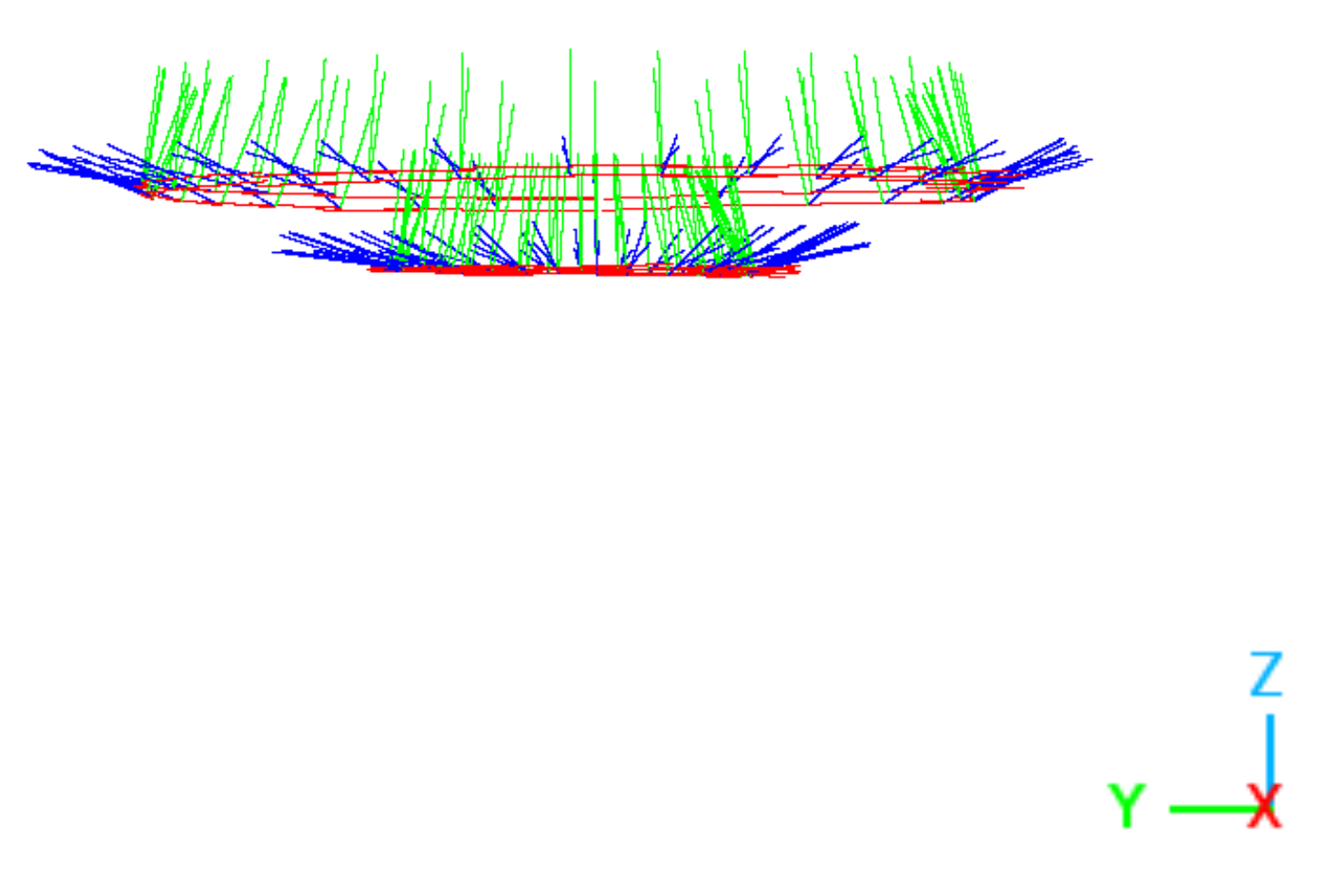}}
\subfigure[]{\label{fig:04_pose_nerfsize}\includegraphics[width=0.23\textwidth]{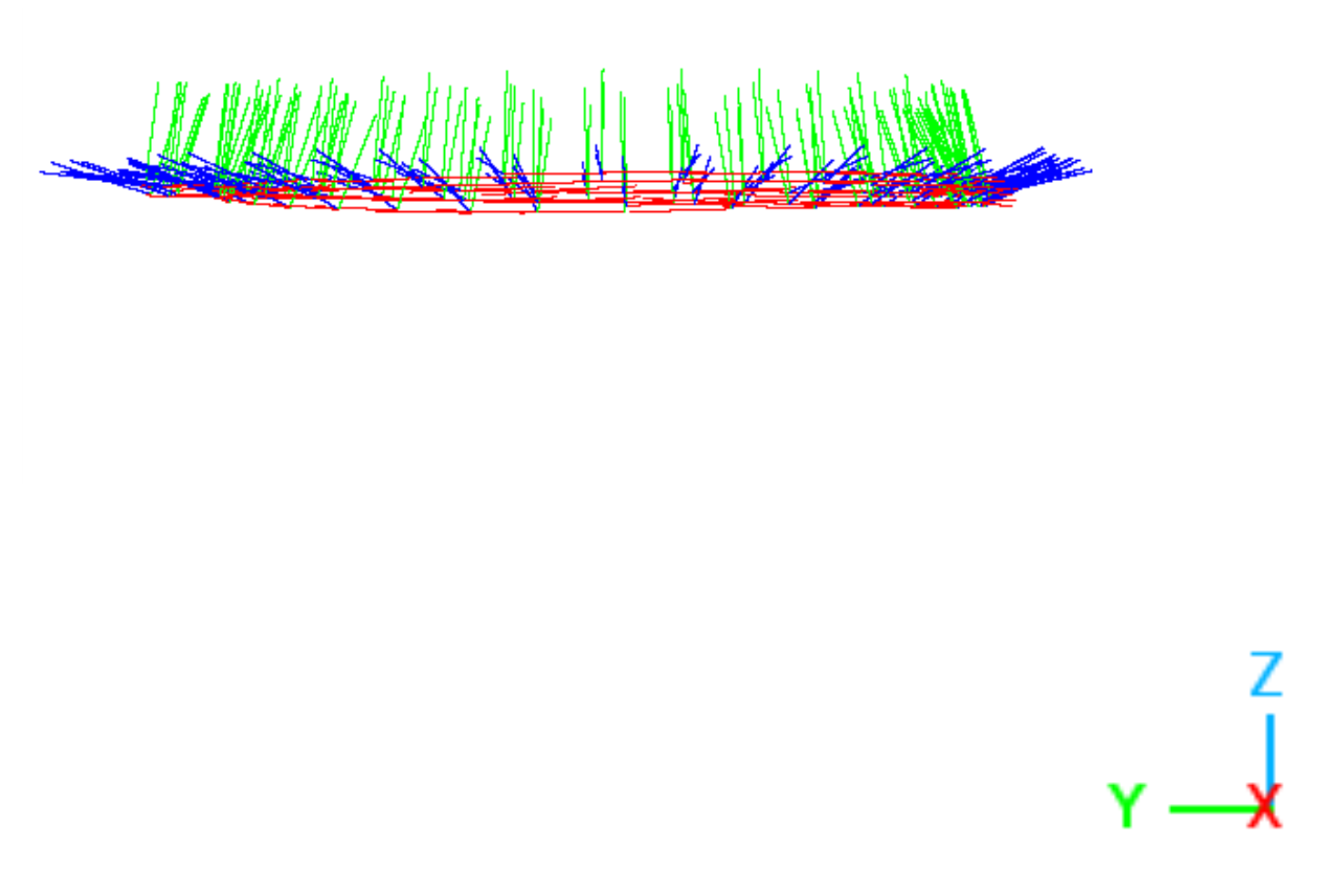}}\vspace{-3mm}
\caption[Camera poses of the trajectory from SfM (trajectory top) via the transformations of Instant NGP implementation versus the internal camera poses from HoloLens (trajectory bottom): \protect{\subref{fig:01_pose_original}} shows the external versus the original internal camera poses, \protect{\subref{fig:02_pose_x-achse}} shows the transformation of the HoloLens camera poses by rotating them 90 degrees around the global x-axis, \protect{\subref{fig:03_pose_attention}} shows the translation to the center of attention and \protect{\subref{fig:04_pose_nerfsize}} shows scaling to the NeRF coordinate system.]{Camera poses for the trajectory from SfM (trajectory top) via the transformations of Instant NGP implementation versus the internal camera poses from HoloLens (trajectory bottom): \protect{\subref{fig:01_pose_original}} shows the external versus the original internal camera poses, \protect{\subref{fig:02_pose_x-achse}} shows the transformation of the HoloLens camera poses by rotating them 90 degrees around the global x-axis, \protect{\subref{fig:03_pose_attention}} shows the translation to the center of attention and \protect{\subref{fig:04_pose_nerfsize}} shows scaling to the NeRF coordinate system.}
\label{fig:Transformations}
\end{figure*}

\paragraph{Transformation of the HoloLens camera poses}
As the first and central step, the HoloLens camera poses have to be transformed into a compatible input format for the NeRF. Such a format is given by the representation as a so-called view matrix $\text{T}_{\text{view}}$, which is a 4$\times$4 transformation matrix in the form of homogeneous coordinates. It consists of translation, rotation and scaling.
The input to the used NeRF follows the OpenGL\footnote{\url{https://learnopengl.com/Getting-started/OpenGL} \\(last access 01/03/2023)} convention, with the camera placed in a right-handed coordinate system, the positive z-axis pointing away from the camera, and the positive x-axis pointing to the right when looking through the camera lens. The y-axes must face the global z-axis, since they correspond to the so-called up-vector. As the z-axis of the HoloLens camera poses $\text{T}_{\text{HoloLens}}$ is orientated in the direction of the z-axis in the global coordinate system, see Figure \ref{fig:01_pose_original}, a transformation by 90 degrees around the global x-axis is required as shown in Figure \ref{fig:02_pose_x-achse} by:

\begin{equation}\label{equ:1}
    \begin{aligned}
     \text{T}_{\text{view}} &= \text{T}_{\text{x,$\alpha$}}  \text{T}_{\text{HoloLens}},
    \end{aligned}
\end{equation}
with 
\begin{equation}
  \begin{aligned}
	\text{T}_{\text{x,$\alpha=90^\circ$}}
	 &= 
     \begin{bmatrix}
     	 1 & 0   & 0  & 0 \\
         0 & \text{cos($\alpha$)} & \text{-sin($\alpha$)}&  0 \\ 
         0 & \text{sin($\alpha$)} &   \text{cos($\alpha$)}& 0 \\ 
         0 & 0 & 0 & 1
     \end{bmatrix} \\
     &=
     \begin{bmatrix}
     	1 & 0   & 0  & 0 \\
         0 & 0 & $-$1 &0 \\ 
         0 & 1 &  0& 0 \\ 
         0 & 0 & 0 & 1
     \end{bmatrix}.
     \end{aligned}     
\end{equation}

Two additional transformations are performed for translation and scaling, based on those in the Instant NGP implementation\footnote{\url{https://github.com/NVlabs/instant-ngp} \\(last access 09/12/2022)}. Firstly, for translation, the transformation matrices are transformed to a common focal point (center of attention), as shown in Figure \ref{fig:03_pose_attention}. For each camera pair, the intersection point between the optical axes is computed, resulting in the focal point. Afterwards the point is subtracted from the current position of the camera, aligning the cameras towards the focal point. This allows the camera poses to be used for the visualization of the object in focus.\newline

Subsequently, scaling is performed on the camera poses to the size of the NeRF coordinate system, see Figure \ref{fig:04_pose_nerfsize}. First, the average distance of all cameras from the origin is calculated by computing the Euclidean norm of the displacement vectors of the camera transformations and summing them. This value is divided by the number of camera transformations to obtain the average distance. The scaling factor is then computed by dividing the camera distances by the average distance and multiplying it with a factor, which is set to 4 by Instant NGP. Final scaling is performed by multiplying each transformation matrix with the final scaling factor.

\paragraph{Structure from Motion}

Structure from Motion (SfM) generally describes the procedure of reconstructing a 3D scene from a set of images taken from different directions and positions. It relies on the calculation and matching of point correspondences within an image sequence from overlapping images by using methods such as SIFT \citep{Lowe}.
In this study, the (incremental) Structure from Motion technique by \citep{Schonberger_2016_CVPR} is used for the external calculation of the camera poses.







\subsection{3D Reconstruction}\label{sec:reconstruction_method}

\paragraph{NeRF}
\vspace{-2mm}
NeRFs enable novel view synthesis of scenes. However, from the perspective of photogrammetry, instead of rendering new 2D views \citep{mueller2019image}, we are interested in the 3D geometry and corresponding color values of the scene.
We consider the density as a kind of pseudo-probability for the occurrence of a surface in 3D space. Considering, positions with high densities indicate a high possibility to be an object point. In the first step, uniform sampling of the density field is achieved by sampling density and color values in the coordinate system of the trained NeRF at equidistant sampling points in a bounding box. In the second step, we filter the positions X = (x,y,z) with high density values using a global threshold $\delta_\text{t}$. Thereby, we assume that object points maintain a higher density  $\delta > \delta_\text{t}$ compared to non-object points.
\newline

\vspace{-2mm}
For the 3D mapping we investigate 3D reconstructions of our scene based on four different types of input data, as shown in Figure \ref{fig:flowchart}. We use the internal HoloLens camera poses and external camera poses calculated by a SfM workflow as described in Section \ref{sec:pose_transform}. In addition to the internal and external camera poses and corresponding images, the pose refinement of the Instant NGP implementation is used. The implementation requires initial poses in order to refine them and is unable to compute poses completely from the scratch. By using the camera pose as an additional variable in the training process, it propagates gradients back onto the camera parameters in order to minimize the loss. \newline


\paragraph{Multi-View Stereo}
In order to compare the reconstructions from NeRFs with a reconstruction from a conventional method, we use a classical Multi-View Stereo (MVS) pipeline \citep{Schonberger_2016_CVPR} on the basis of the output of the Structure from Motion in Section \ref{sec:pose_transform}. Accordingly, the same HoloLens camera poses as for the NeRFs serve as input here, which makes the reconstructions comparable in the same coordinate system. 
On the one hand, we use the output of SfM for a sparse reconstruction. On the other, hand we generate a dense reconstruction with MVS. 
MVS \citep{Schonberger_MVS} takes the information from the sparse model from SfM to for pixelwise computation of depth information in an image.

\section{Dataset}\label{dataset}
\begin{figure}[h!]
\begin{center}
		\includegraphics[width=0.98\columnwidth]{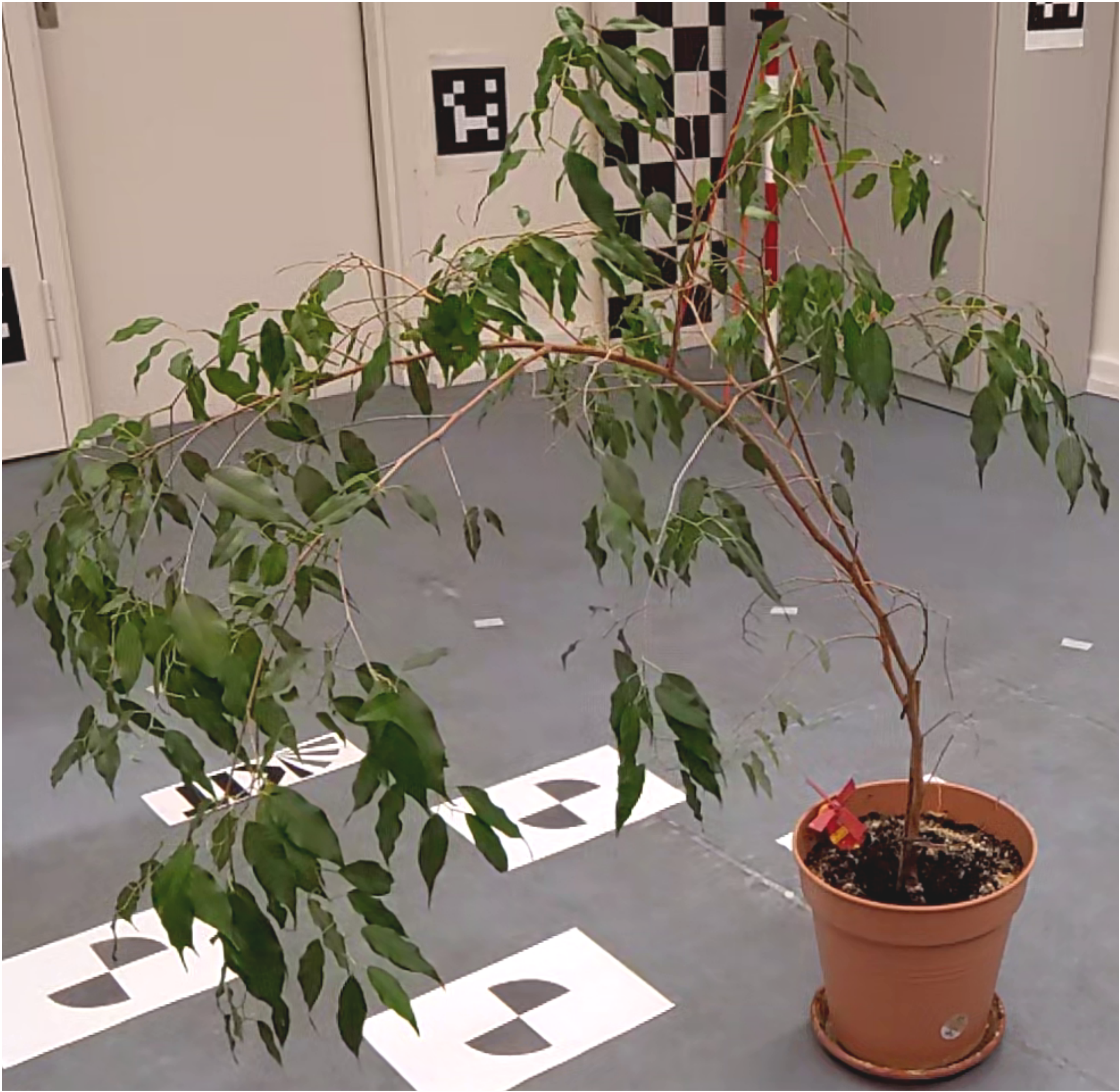}
		\vspace{-3mm}
	\caption{Visualization of an image of the captured Ficus plant as our measurement object using the Microsoft HoloLens RGB camera.}
\label{fig:ficus}
\end{center}
\end{figure}

Our expertiments are based on a dataset captured by the Microsoft HoloLens, which includes an indoor scene of a plant (Ficus) on a plane surface, see Figure \ref{fig:ficus}. 
The HoloLens provides an interesting interface for the NeRF, as it generates the required input data, camera poses and associated sensor images. In general, HoloLens, developed by Microsoft and firstly released in 2018, embodies the world's first fully autonomous holographic computer and has become an important device for all kinds of applications, such as 3D mapping and modelling of indoor scenes (\citeauthor{jaeger}, \citeyear{jaeger}; \citeauthor{hololens_weinmann}, \citeyear{hololens_weinmann}).

\begin{figure*}[h!]
\begin{center}
		\includegraphics[width=0.76\textwidth]{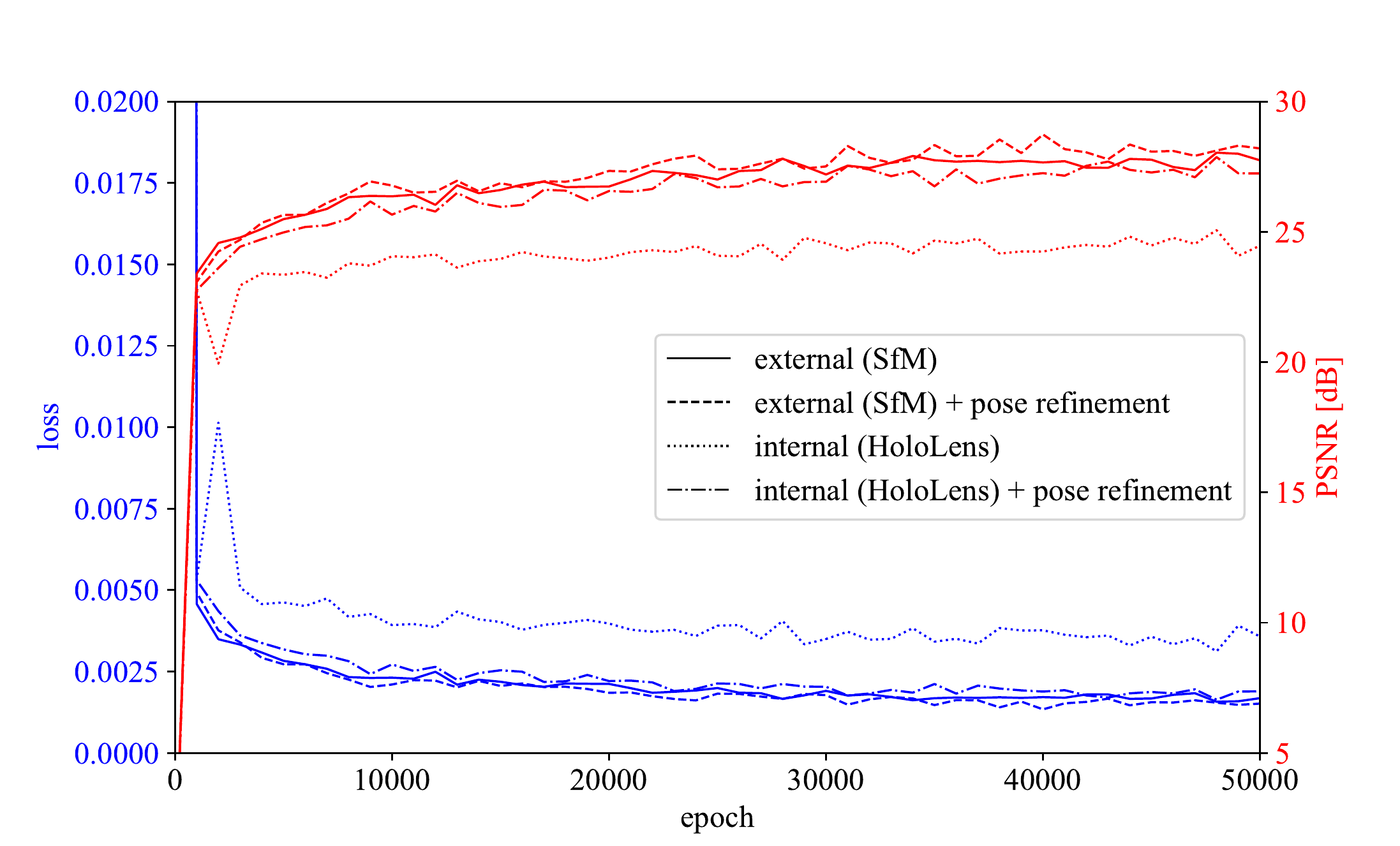}
	\caption{Comparison of the Peak-Signal-to-Noise-Ratio (PSNR) in \si{\dB} $\uparrow$ and loss $\downarrow$ during the training processes. The red curves show the PSNR, the blue curves the loss. The HoloLens images with internal HoloLens camera poses, internal HoloLens camera poses with pose refinement, external (SfM) camera poses, and external (SfM) camera poses with pose refinement are considered.}
\label{fig:training}
\end{center}
\end{figure*}

\begin{figure*}[h!]
	\centering
\subfigure[]{\label{fig:Ficus_HoloLens_intern}
	\includegraphics[width=0.9\columnwidth]{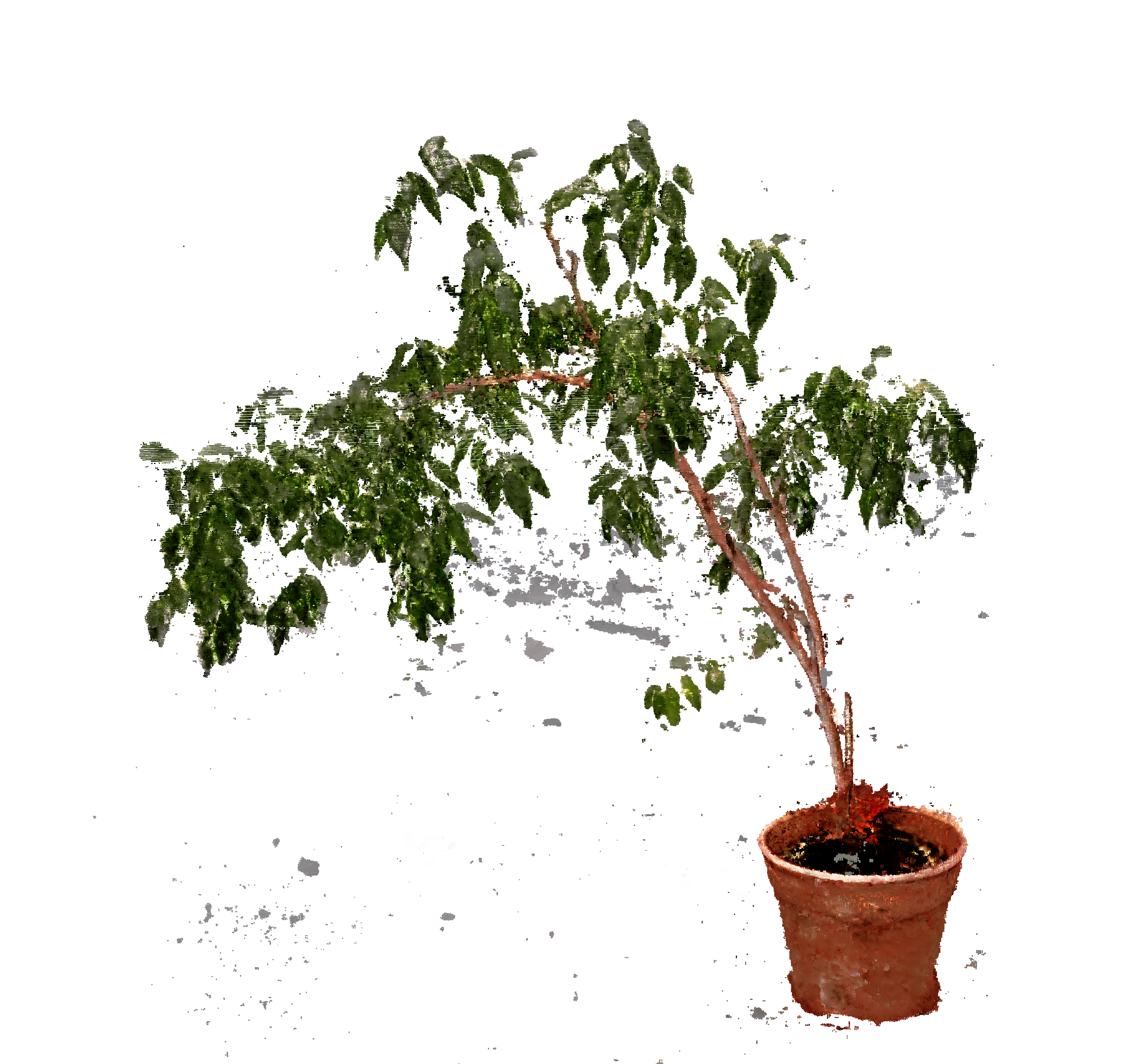}}   
\subfigure[]{\label{fig:Ficus_HoloLens_intern_pose}  
     \includegraphics[width=0.9\columnwidth]{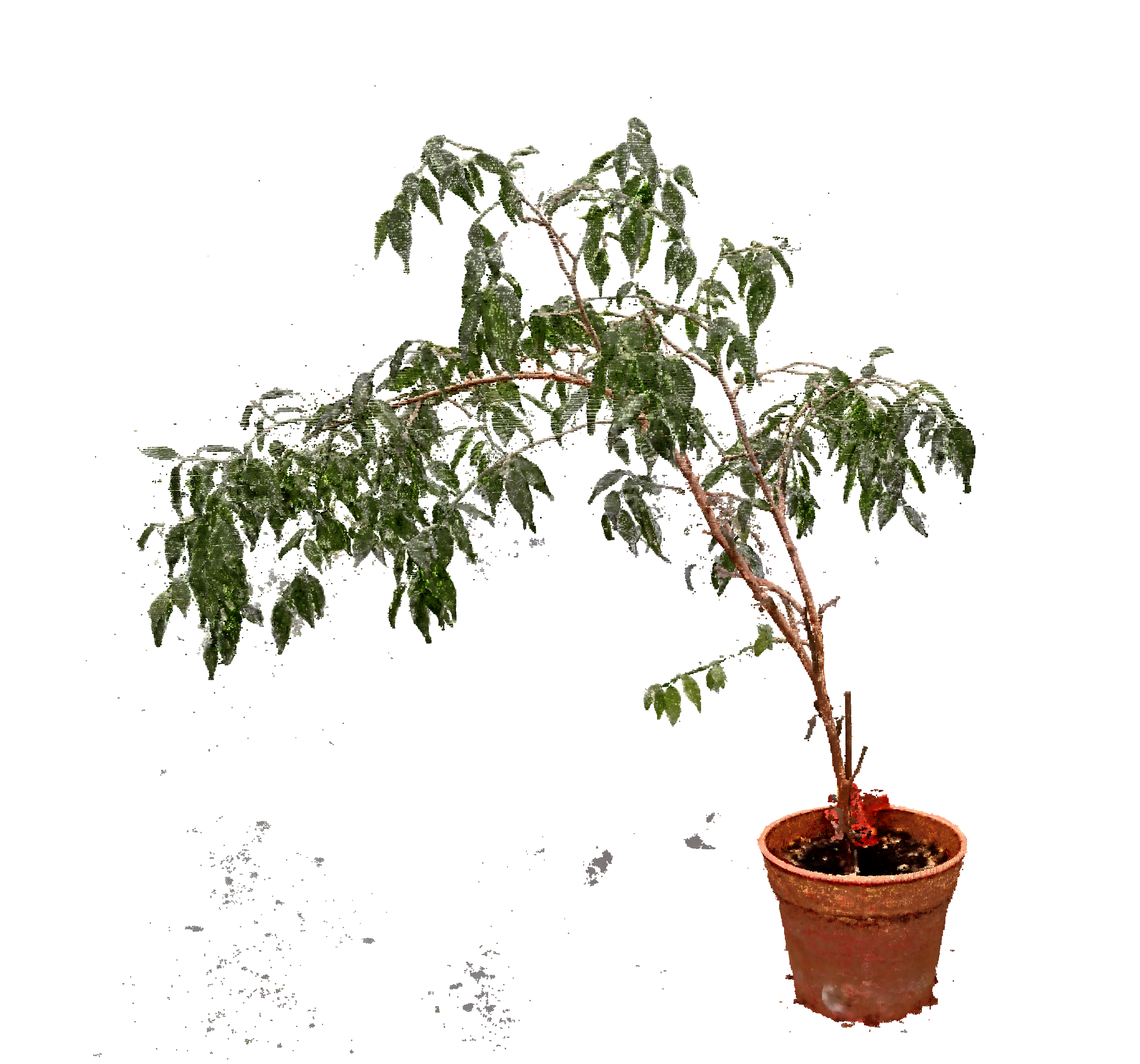}}\\   
\subfigure[]{\label{fig:Ficus_HoloLens_COLMAP}
	\includegraphics[width=0.9\columnwidth]{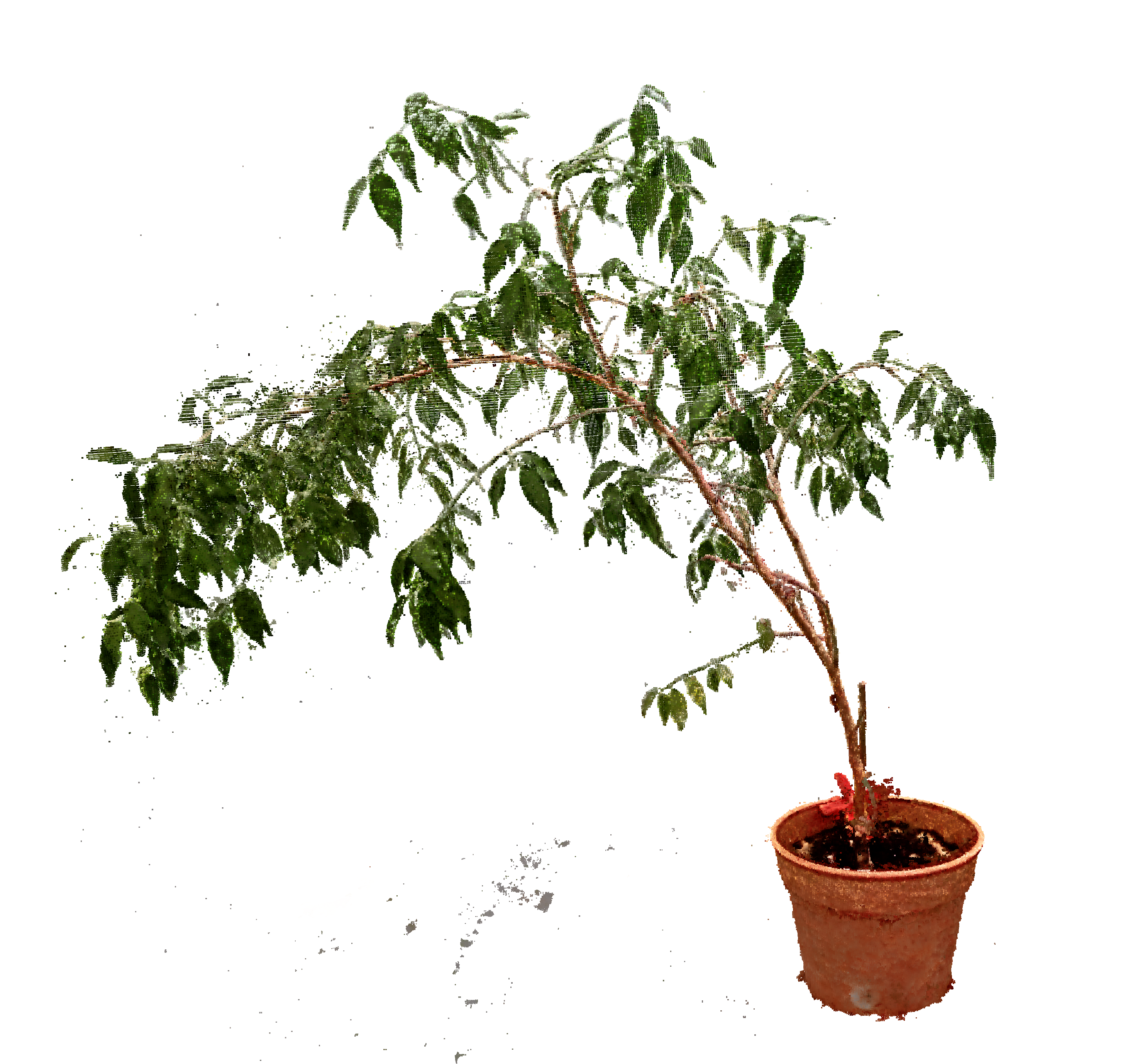}}    
\subfigure[]{\label{fig:Ficus_HoloLens_COLMAP_pose}  
     \includegraphics[width=0.9\columnwidth]{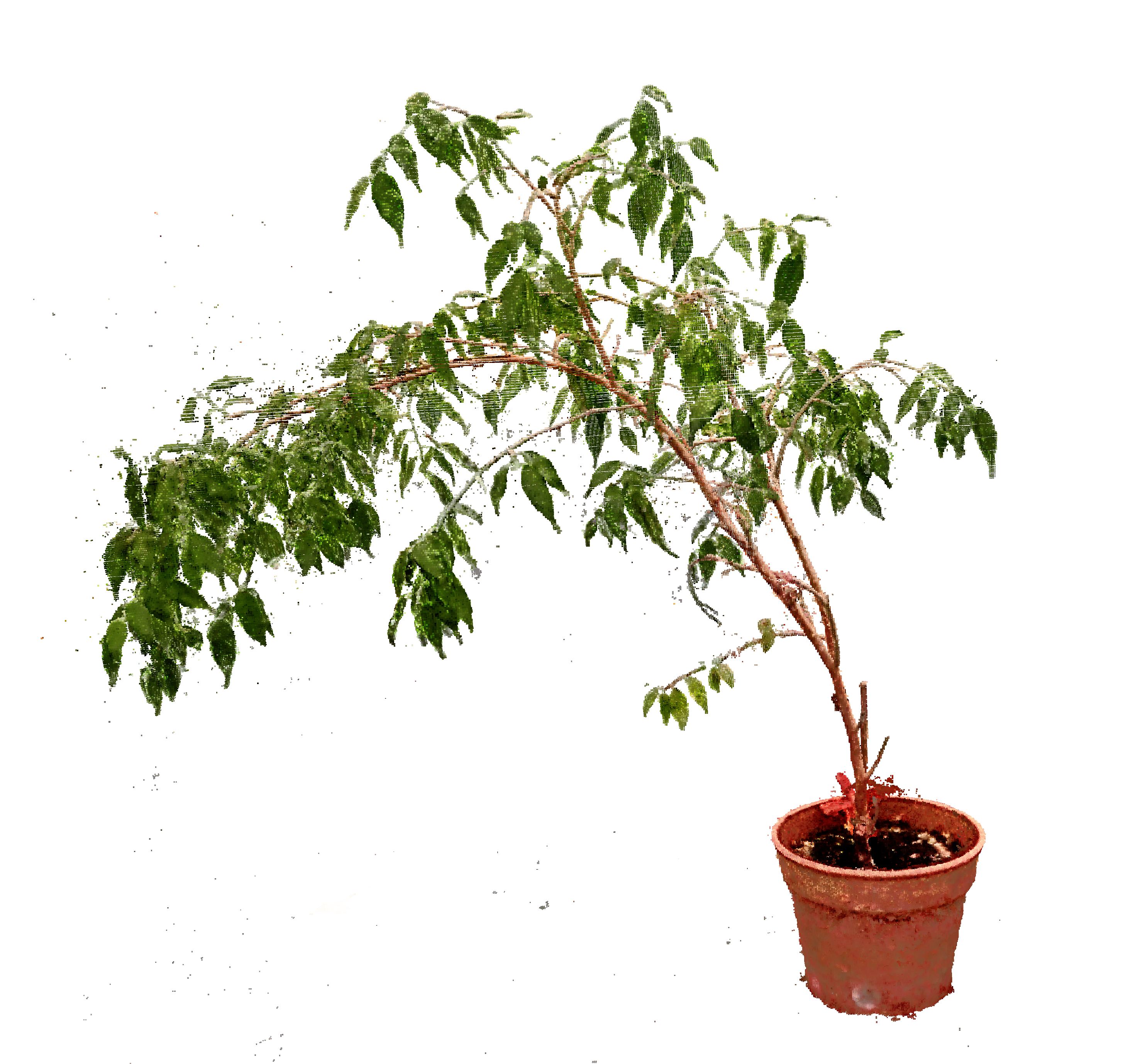}}   \\ 
     \subfigure[]{\label{fig:Ficus_HoloLens_COLMAP_MVS_sparse}
	\includegraphics[width=0.9\columnwidth]{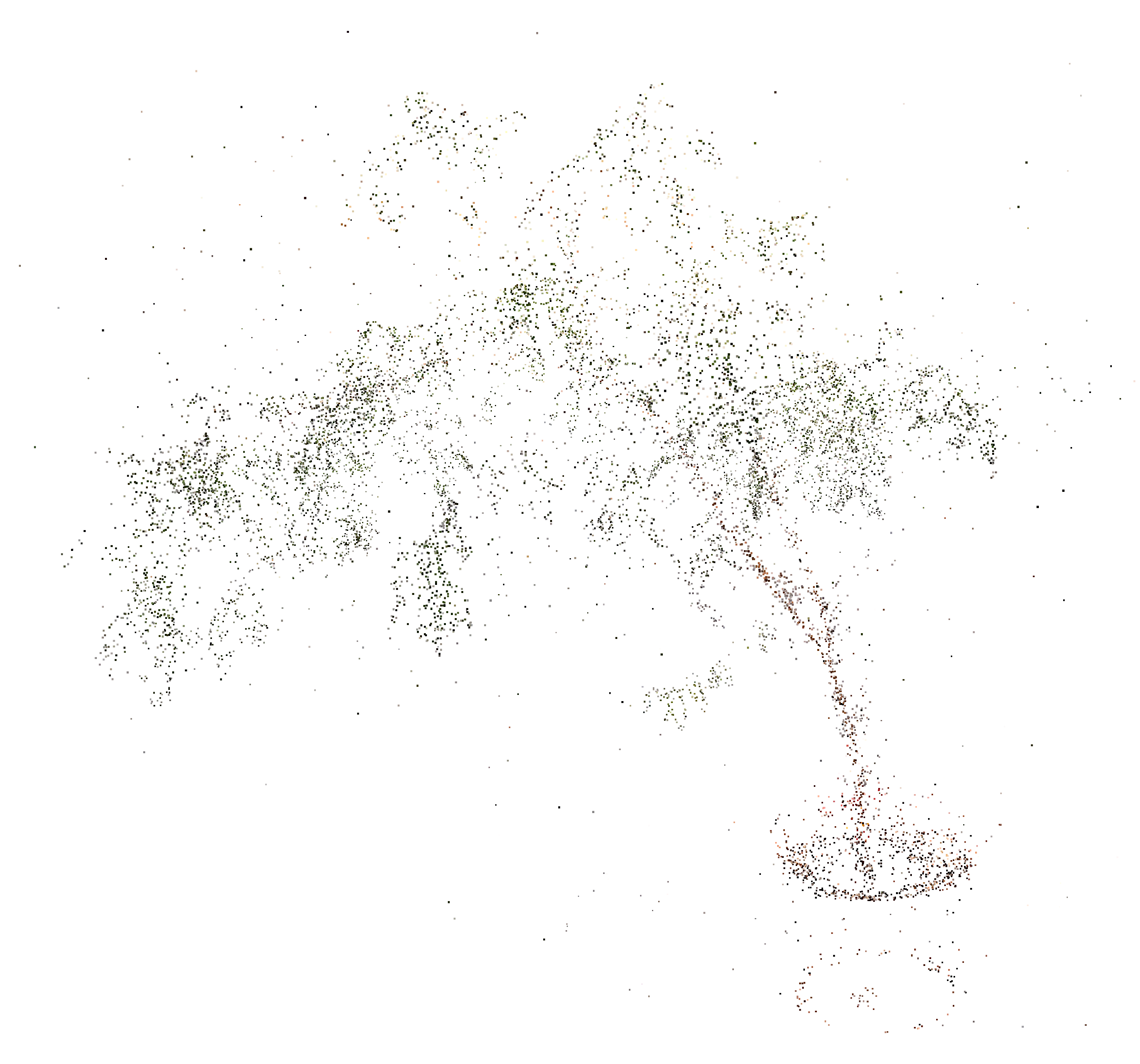}}   
\subfigure[]{\label{fig:Ficus_HoloLens_COLMAP_MVS_dense}
	\includegraphics[width=0.9\columnwidth]{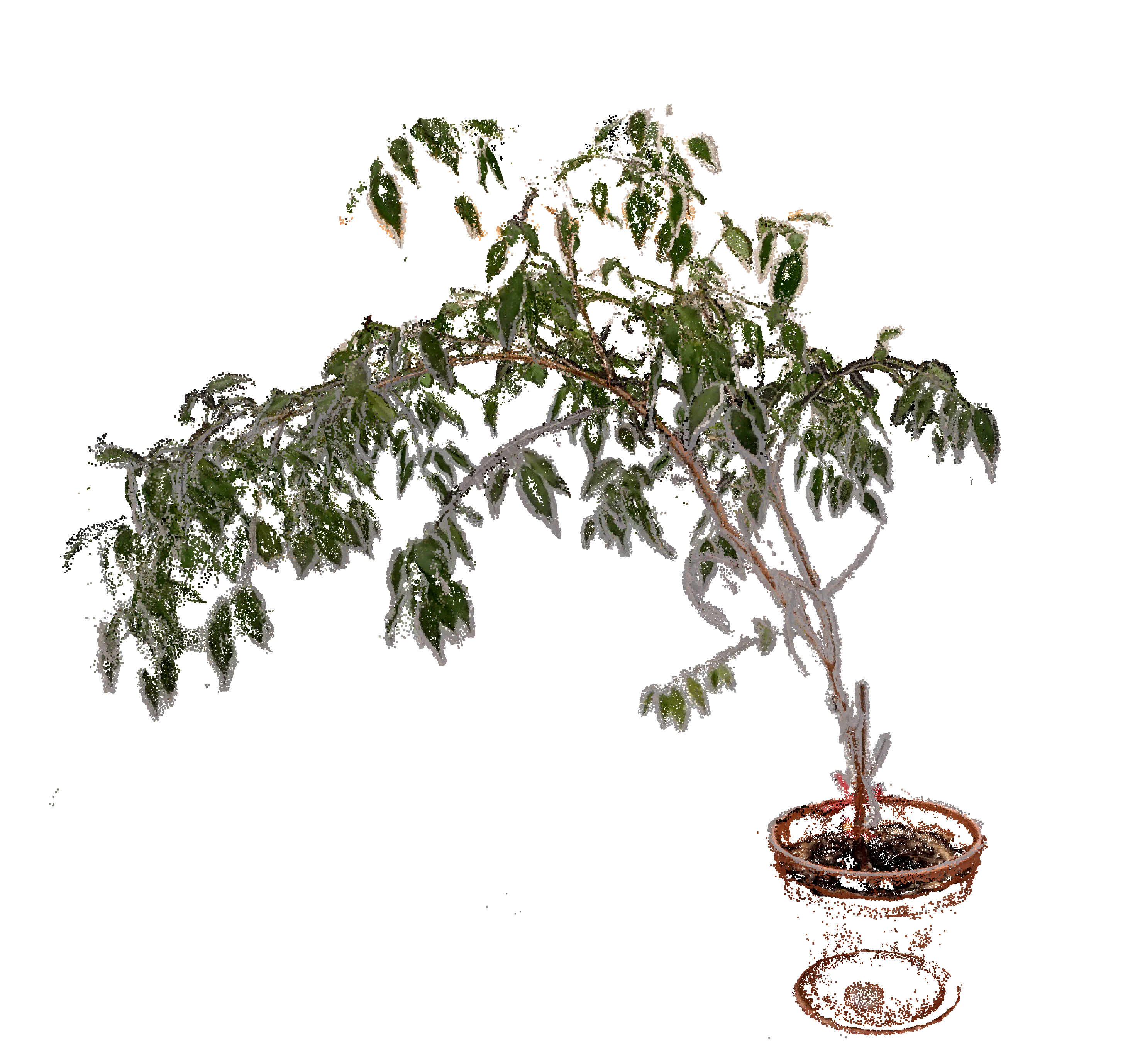}}    
	\caption[Comparison of the 3D reconstructions from NeRFs using a global density threshold $\delta_{\text{t=15}}$. For HoloLens images and \protect{\subref{fig:Ficus_HoloLens_intern}} internal camera poses, \protect{\subref{fig:Ficus_HoloLens_intern_pose}} with pose refinement and \protect{\subref{fig:Ficus_HoloLens_COLMAP}} external camera poses, \protect{\subref{fig:Ficus_HoloLens_COLMAP_pose}} with pose refinement. Compared to the \protect{\subref{fig:Ficus_HoloLens_COLMAP_MVS_sparse}} sparse and \protect{\subref{fig:Ficus_HoloLens_COLMAP_MVS_dense}} dense point cloud from external camera poses with MVS.]
	{Comparison of the 3D reconstructions from NeRFs using a global density threshold $\delta_{\text{t=15}}$. For HoloLens images and \protect{\subref{fig:Ficus_HoloLens_intern}} internal camera poses, \protect{\subref{fig:Ficus_HoloLens_intern_pose}} with pose refinement and \protect{\subref{fig:Ficus_HoloLens_COLMAP}} external camera poses, \protect{\subref{fig:Ficus_HoloLens_COLMAP_pose}} with pose refinement. Compared to the \protect{\subref{fig:Ficus_HoloLens_COLMAP_MVS_sparse}} sparse and \protect{\subref{fig:Ficus_HoloLens_COLMAP_MVS_dense}} dense point cloud from external camera poses with MVS.}
\label{fig:pointclouds_COLMAP_SLAM}
\end{figure*}
\newpage
HoloLens\footnote{\url{https://www.microsoft.com/en-us/hololens/hardware} \\(last access 02/20/2023)} generation 2 was released in 2019 and features improved camera technology compared to the first generation such as higher resolution and better color depth, resulting in sharper and more detailed images.\newline 

The HoloLens 2 server application \citep{HoloLens2_Streaming} is used for requesting the data in the HoloLens. The system provides access to all the HoloLens 2 sensors, including the images from the 1920\,$\times$\,1080 photo/video RGB camera and corresponding camera pose of the device in 3D space. In addition, device calibration data can be retrieved by the internal orientation (camera intrinsics).
The HoloLens images and corresponding camera poses were captured with a hemispherical camera framing. Thereby step sizes of 32 scanning points at a height of about 120\,cm, with two different viewing angles have been employed, which results in a total of 64 images.

\section{EXPERIMENTS AND RESULTS}\label{RESULTS}
In this section, we present our experiments and results by a quantitative evaluation on analyzing the training process in Section \ref{sec:training}. This is followed by a qualitative analysis in Section \ref{sec:reconstruction} of the resulting 3D reconstructions. We investigate the impact of the camera poses in general, the comparison of point clouds from NeRFs trained with different input sets based on HoloLens data in Section \ref{dataset} as well as photogrammetric reconstructions.

\subsection{Training}\label{sec:training}

The training process of the NeRFs proceeds differently based on the chosen configurations, as shown in Figure \ref{fig:training}. In particular, we use the Peak-Signal-to-Noise-Ratio (PSNR) in \si{\dB} between the input RGB images of the training data and the rendered images for the accuracy measurement while training.
Comparing the training, both the internal HoloLens and the external SfM camera poses lead to a convergence of the NeRF. This occurs at approximately 20,000 training epochs, which corresponds to a duration of between 2 and 5\,min training. 
The internal HoloLens camera poses achieve best results of about 25\,dB. In contrast, more than 27\,dB can be achieved with the external camera poses from SfM in pre-processing. Remarkably, pose refinement by training the extrinsics can increase the performance for the internal camera poses to a comparable level of about 27\,dB. However, no further increase in PSNR is achieved for the external camera poses by pose refinement. 
For all configurations, the loss behaves inversely proportional. Based on these training results, conclusions can be drawn about the relative precisions of the different type of camera poses.




\subsection{3D Reconstruction}\label{sec:reconstruction}

Finally, Figure \ref{fig:pointclouds_COLMAP_SLAM} compares the 3D reconstructions from the NeRFs trained on different input data by using a global threshold $\delta_{\text{t=15}}$, and the sparse and dense point clouds. In general, 3D reconstructions from NeRF on HoloLens data can be generated directly with the internal camera poses as well as from the external camera poses calculated via SfM.\newline

The visual quality of the reconstructions corresponds to the achieved PSNR values in Figure \ref{fig:training}.
This is particularly evident from the artifacts in the reconstruction from HoloLens internal camera poses with no pose refinement in Figure \ref{fig:Ficus_HoloLens_intern}. The training course also shows a lower maximum PSNR of 25\,dB compared to those of the other three training processes with PSNR values of 27\,dB. In this case, artifacts are located in empty space. This effect rarely occurs during 3D reconstruction based on the internal camera poses with pose refinement, as Figure \ref{fig:Ficus_HoloLens_intern_pose} shows.
The external camera poses provide adequate qualitative reconstruction results without in Figure \ref{fig:Ficus_HoloLens_COLMAP} and with pose refinement in Figure \ref{fig:Ficus_HoloLens_COLMAP_pose}. Only small artifacts disappear with pose refinement. Overall, all input data provide sufficient 3D reconstructions from NeRFs with minor color differences. In particular, the surface of the pot of the plant can be reconstructed well using NeRFs. \newline 

In contrast, the MVS approach does not provide a complete reconstruction of the object, which is especially noticeable on the pot. This occurs for both the sparse reconstruction in Figure \ref{fig:Ficus_HoloLens_COLMAP_MVS_sparse} and dense reconstruction in Figure \ref{fig:Ficus_HoloLens_COLMAP_MVS_dense}. Additionally, the dense point cloud contains gray artifacts at the fine structures of the branches.


\section{DISCUSSION}\label{DISCUSSION}

This research investigates the application of camera poses from Microsoft HoloLens trajectories for 3D reconstruction. On the one hand, internal camera poses directly retrieved from the HoloLens trajectory via a server application have been investigated. On the other hand, external camera poses that were calculated in the conventional manner via Structure from Motion were considered. For both scenarios, an enhanced pose refinement was additionally applied by training the camera extrinsics. \newline
 

It could be demonstrated that, after a simple rotation around the x-axis, the internal HoloLens camera poses are sufficient for NeRF convergence in approximately 20,000 training epochs. This enables a 3D reconstruction using NeRF coordinates by sampling. Four investigations are considered as input for the corresponding images: The internal HoloLens camera poses, external camera poses from SfM, both with and without pose refinement. 
Overall, the results show varying quantitative and qualitative performance in training and 3D reconstruction based on the utilized camera poses.
Considering the training process the unrefined internal HoloLens camera poses provide PSNR of about 25\,dB. With pose refinement of the internal camera poses, the training process improves to about 27\,dB. This is comparable with the external camera poses from SfM, which achieve higher PSNR values of around 27\,dB, both unrefined and refined.
We assume that improved poses lead to a superior training process in terms of the PSNR values and consequently better 3D reconstructions, which is confirmed by the qualitative results. Thereby the unrefined internal HoloLens camera poses contains more huge artifacts in the 3D reconstruction. However, by pose refinement of the internal camera poses, the artifacts are reduced, and the reconstruction is comparable to the reconstruction from external calculated camera poses. Each external camera poses without and with pose refinement, contain only a few small artifacts. We suggest that the externally calculated poses are already quite accurate and therefore do not improve further with pose refinement.\newline

Nevertheless, the results from 3D mapping using NeRF are notably superior to the classical photogrammetric method of dense Multi-View stereo (MVS) reconstruction from camera poses via SfM for our dataset. 
The NeRF reconstructions yield better results on untextured, homogeneous surfaces. This is especially evident for the pot of the plant, which apparently fails to be reconstructed with the conventional MVS. In addition, fine structures in MVS reconstruction contain gray artifacts, as can be seen in the branches of the plant and an inferior level of detail. Some color differences within the NeRF reconstructions are caused by the directionality of color in the NeRFs, as opposed to density. However, the color differences are minor and do not harm the overall impression of the reconstruction.

\section{CONCLUSION}\label{CONSLUSION}

In this paper, we presented a workflow for the extraction of high resolution 3D reconstructions almost directly from Microsoft HoloLens data under the application of Neural Radiance Fields (NeRFs). Thereby, the impact of the camera poses has been investigated using a quantitative analysis by considering the training process, as well as a qualitative analysis by regarding the final 3D reconstructions.\newline

We demonstrated that the internal HoloLens camera poses und corresponding images as input data are able to provide convergence of the NeRF during training. This enables the generation of a 3D reconstruction from positions with high density values in the NeRF coordinate system. Improvements in the training process and resulting 3D reconstruction can be achieved by pose refinement while training the NeRF. This enables a comparable quality in the training process and resulting point cloud as achieved by external camera poses calculated in pre-processing using approaches such as Structure from Motion. It demonstrates the impact of the camera poses on the quality of the 3D reconstruction. In addition, among all pose investigations, the NeRF reconstructions outperform the conventional photogrammetric method using Multi-View Stereo.\newline

In summary, the combination of internal HoloLens camera poses and associated images with NeRFs offers an immense potential for enabling highly detailed, colored, mobile 3D mappings of a scene in a straightforward workflow. 
In future work, we suggest using a 3D region growing algorithm instead of a global density threshold in terms of artifacts removal, assuming that all object points in the scene are spatially connected.

\bibliographystyle{elsarticle-harv} 
\bibliography{ISPRSguidelines_authors}




\end{document}